# Mero Nagarikta: Advanced Nepali Citizenship Data Extractor with Deep Learning-Powered Text Detection and OCR

Sisir Dhakal[1], Sujan Sigdel[2], Sandesh Prasad Paudel[3], Sharad Kumar Ranabhat[4], Nabin Lamichhane[5]

[1,2,3,4,5]Department of Electronics and Computer Engineering, Institute of Engineering (IOE) Pashchimanchal Campus, Tribhuvan University, Pokhara, Gandaki, Nepal

*Abstract -* *Transforming text-based identity documents, such as Nepali citizenship cards, into a structured digital format poses several challenges due to the distinct characteristics of the Nepali script and minor variations in print alignment and contrast across different cards. This work proposes a robust system using YOLOv8 for accurate text object detection and an OCR algorithm based on Optimized PyTesseract. The system, implemented within the context of a mobile application, allows for the automated extraction of important textual information from both the front and the back side of Nepali citizenship cards, including names, citizenship numbers, and dates of birth. The final YOLOv8 model was accurate, with a mean average precision of 99.1% for text detection on the front and 96.1% on the back. The tested PyTesseract optimized for Nepali characters outperformed the standard OCR regarding flexibility and accuracy, extracting text from images with clean and noisy backgrounds and various contrasts. Using preprocessing steps such as converting the images into grayscale, removing noise from the images, and detecting edges further improved the system's OCR accuracy, even for low-quality photos. This work expands the current body of research in multilingual OCR and document analysis, especially for low-resource languages such as Nepali. It emphasizes the effectiveness of combining the latest object detection framework with OCR models that have been fine-tuned for practical applications.*

## 1. Introduction

In today's digital world, the management of personal data has become increasingly important, particularly when it comes to identity documents such as citizenship cards. Back in Nepal, these papers are used for almost every transaction and any kind of legal process, but the users always have to take the pain of writing down the important information from these papers. This manual entry of information is not only very time-consuming, but it is also prone to human error, which can, in turn, cause problems in situations where precise data is of the utmost importance. It is obvious that a system is needed to streamline the process of collecting and managing all of this data.

The newest developments in deep learning, especially in Convolutional Neural Networks (CNNs) [1] and object detection methods, however, provide hopeful answers to this dilemma. YOLO (You Only Look Once) and other technologies have changed the game with scene text detection, and it's possible to pull text from an image with very high accuracy and speed.

Latha et al. [2], in their research, utilized the YOLO model on the COCO 'Common Objects in Context' dataset for real-time multilingual scene text detection and addressed the challenges caused by varying text characteristics and backgrounds of the image. They also highlighted the effectiveness of the convolutional neural network(CNN) in processing and effectively identifying the scene text in various contexts. Similarly, in 2024, Liu et al. [3] presented YOLOv5ST, a lightweight model fine-tuned specifically for fast scene text detection with increased speed and accuracy in inference. Their experiment showed an impressive 26% increase in speed during inference compared to the baseline model. In 2023, Gao and Liu [4] proposed Invo-YOLOv5, an enhanced deep learning-based text detection algorithm that replaces traditional convolution layers with inverse convolution in YOLOv5. Their study demonstrated an improved efficiency in the detection of text by employing K-means clustering to optimize the anchor frames in diverse text sizes, which they validated through CRNN and Tesseract OCR on complex class samples.

El Abbadi et al. [5], through their paper, proposed a scene text recognition system for enhanced feature extraction based on YOLOv5 and Maximally Stable Extremal Regions (MSERs). They reported impressive performance metrics for their system, achieving recall of 96%, precision of 80%, and an F1 score of 87.6%. Chaitra et al. [6] studied the influence of applying YOLOv5 toward the enhancement of text detection and recognition implemented within video frames, including the use of TesseractOCR for enhancing text extraction, specifically in complex environments.

The main purpose of this paper is to actually experiment and see how well the YOLOv8 [7] does on text detection and utilize it along with optimized Pytesseract OCR to extract text from Nepali citizen's ID cards. The results shown by the model were pretty impressive, and through its evaluation, we aimed to develop a stable and efficient system for easy retrieval of users' most important information and ultimately improve the document management process for the user. This research can be used as a base for future works in optimizing text recognition for the Nepali language, and hopefully, it will inspire other researchers to do more work in improving accuracy and efficiency. The results of this study will help to further the field of multilingual text detection as a whole and lead to technological strides that include everyone.

In this project, we first briefly describe the related works and the current approaches for scene text detection and recognition. In section 2, we discuss specific information about the dataset and the preprocessing stage. Section 3 discusses the evaluation and performance metrics. We then outline the results of the experiments that were carried out (Section 4). We analyze the results in the last section (Section 5) and recommend further studies.

## 2. Methodology

### 2.1. Dataset Interpretation

#### 2.1.1. Collection

For this project, the four members of the group contacted their closest friends and asked for copies of their citizenship cards, while also assuring them of confidentiality through a Non-Disclosure Agreement (NDA) to prevent misuse. Some of the images were captured using our own camera to ensure clarity. Subsequently, all these pictures were compiled in Google Drive and organized into folders to establish a dataset with diverse variations. We collected 250 images with different text types, formats, and lighting conditions.

#### 2.1.2. Description

The dataset includes pictures of the front and back of the Nepali citizenship card. We took photos of some citizenship cards while they were still enclosed in transparent plastic cases and captured others without the case. This approach gave the model more diversity in clarity, glare, and reflections, addressing real-world situations. Also, the pictures of the card were taken in different lighting, angles, and other atmospheric conditions to increase the variance in the dataset.

### 2.2. Data Preprocessing

This project followed the following steps for the preprocessing of the data:

#### 2.2.1. Data Organization

The photos of the citizenship cards were carefully collected, archived, and then organized into folders. During this process, duplicate photos were discovered which were later removed from the dataset.

### 2.2.2. *Data Normalization*

The pixel intensities of each photo were normalized to a standard range between 0 and 1. This helped us speed up and stabilize the learning process by ensuring that the model processed each image uniformly, regardless of differences in quality between images within the dataset [8].

### 2.2.3. *LabelImg*

We used an open-source labeling tool called LabelImg to facilitate the labeling process for the front and back sides of the citizenship images [9]. We surrounded the areas containing the relevant text with rectangles and made annotations in YOLO format. Each annotation file included class labels followed by the coordinates of the top-left corner and the dimensions to the bottom-right corner for each image. We saved the files in text format corresponding to each image.

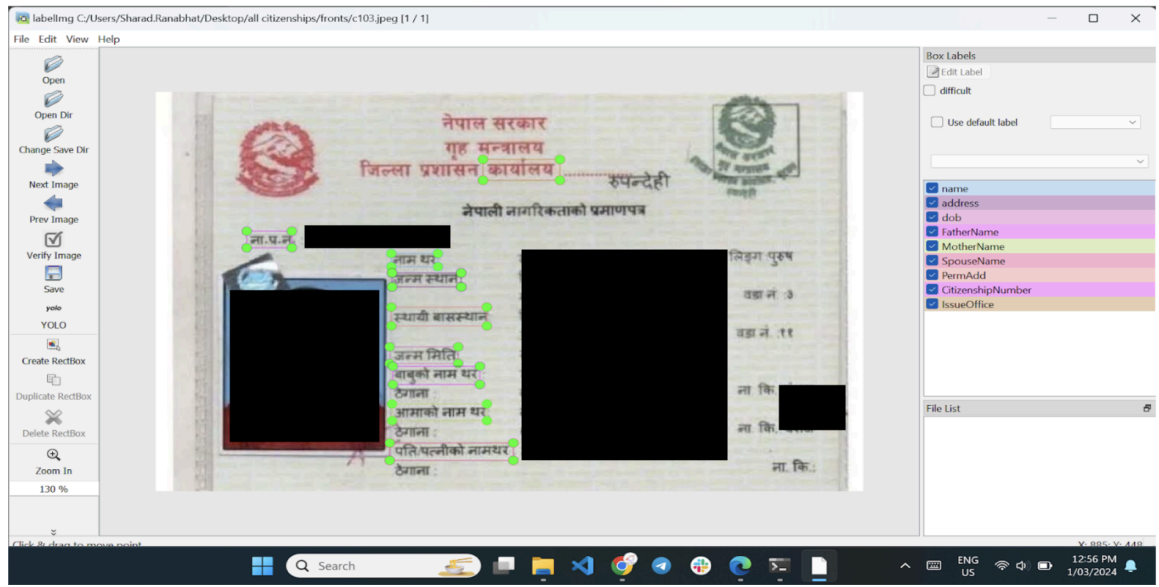

**Figure 1:** Creating annotations of regions of interest (ROIs) using LabelImg. The black patch appears only in this paper to conceal sensitive citizenship information; it does not exist in the original image used for labeling and model training.

### 2.2.4. *Data Splitting*

We divided the datasets into 210 instances of preprocessed citizenship images for training and 40 for validation. We used the popular Python package split-folders inorder to shuffle and split the dataset while also preserving the original distribution [10]. This division ensured enough data for both the training and during the testing of model's performance [11].

### 2.2.5. *Data Augmentation*

Data Augmentation reduces overfitting when it is used with pre-trained models. It makes the model aware of a more comprehensive array of features in the images. Similarly, the augmented dataset also increases the robustness of the model when fine-tuning for specific objectives such as object detection and text recognition. So, to increase the number of images and variability in the augmented dataset, we used image augmentation [12]. The various manipulation techniques include basic operations like rotation or flipping of the image, scaling of the image, and the image's brightness and contrast. It is beneficial for transfer learning as it enhances the size of the dataset.

### 2.2.6. *Creating YOLO Configuration Files*

After that, we defined the YOLO configuration file (.yaml file), which consists of the model definition, training procedure, and paths to the training and validation datasets. In YOLO, this configuration file is crucial for initializing the architecture, the number of classes, class labels, and other vital parameters when training and optimizing the model.

### 2.3. YOLOv8 and Transfer Learning

This project used transfer learning to efficiently adapt a YOLOv8 model for training on our custom dataset. Transfer learning is a technique whereby a model trained on some significant data, such as the COCO set, is fine-tuned on a specific data set [13]. This approach allowed us to take advantage of the knowledge already embedded in YOLOv8 and train the model quickly.

The new YOLOv8 [7] model, which builds on previous YOLO versions represents the cutting edge in object detection, classification, and segmentation tasks. This version improves both speed and accuracy through architectural improvements. The model incorporates a CSP-based Backbone, which efficiently processes feature maps, and a robust Head for precise object localization and classification. By utilizing this pre-trained model and fine-tuning it on our dataset, we achieved high performance while minimizing training time and computational costs.

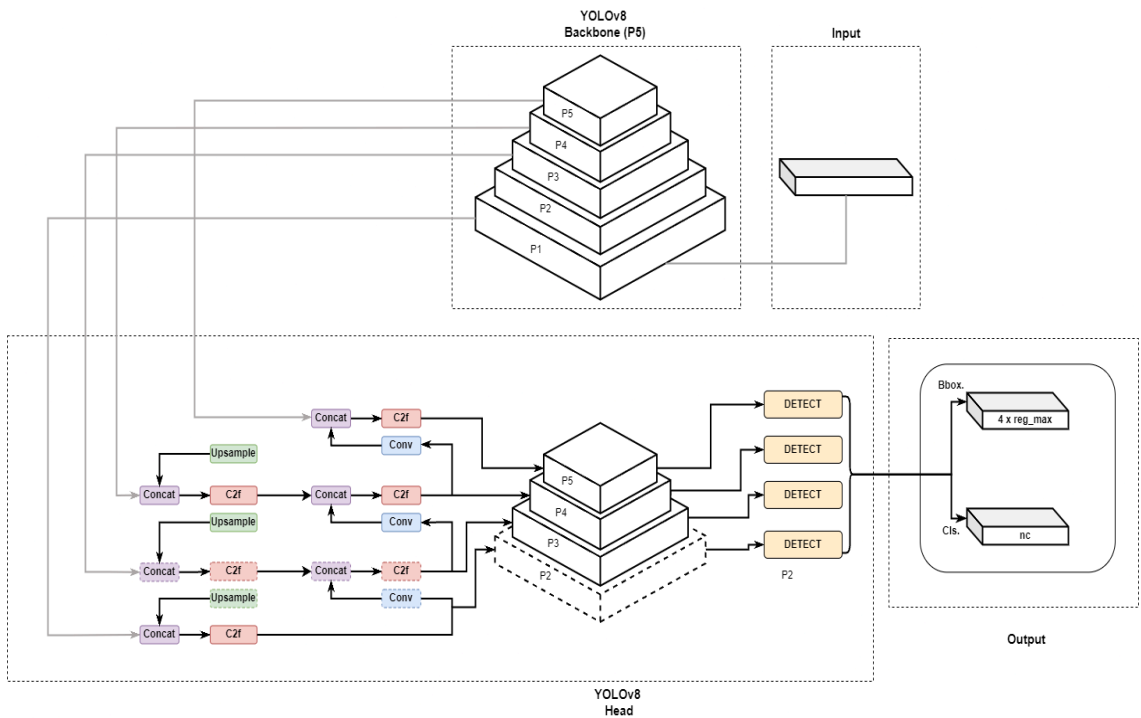

Figure 2: Reference image for YOLOv8 architecture [14].

### 2.4. Model Building

We used YOLOv8 along with it's pretrained weights and added fully connected layers followed by a softmax layer and used transfer learning to develop two models each for front and back sides of citizenship to detect the desired classes. This dual-model approach covers all relevant information, improving the accuracy of our detection system.

### 2.5. Hyperparameters

In order to get the best results, we experimented with different sets of hyperparameter as hyperparameters are crucial in learning and generalizing the model.

Final key hyperparameters and their configurations include:

| Hyperparameters | Front Model | Back Model |
| --- | --- | --- |
| Image Size | 640 * 640 | 640 * 640 |
| Learning Rate | Adaptive | Adaptive |
| Batch Size | 08 | 04 |
| Epochs | 100 | 50 |
| Optimizers | {Adam, AdamW} | {Adam, AdamW} |

Table 1: Hyperparameters.

In this project, we used a combination of loss functions for classification and bounding box regression to train the models for multiple-class detection effectively.

### 2.5.1. Categorical Cross-Entropy (CCE) for Classification:

Since we dealt with multiple classes (different labels on the citizenship card), we used Categorical Cross-Entropy (CCE) as the primary loss function for classification. CCE calculates the dissimilarity between the predicted class probabilities and the actual one-hot encoded ground truth labels for multiple classes.

The formula for Categorical Cross-Entropy is as follows:

$$Categorical\ Entropy\ =\ -\sum_{neuron=1}^{classes} y_{true_{neuron}} * ln\left(y_{pred_{neuron}}\right) \quad (1)$$

This loss encourages the model to maximize the likelihood of the correct class for each label on the citizenship card.

### 2.5.2. DFL (Distribution Focal Loss) for Bounding Box Regression:

This project utilizes Distribution Focal Loss (DFL) to refine the object localization. DFL helps to predict the boundaries more accurately by applying a higher penalty to errors on smaller and more difficult-to-detect objects, focusing the model's attention on more challenging regions.

### 2.5.3. CIoU (Complete Intersection over Union) for Bounding Box Regression:

CIoU is a specialized loss function which is used to measure how well the predicted bounding box overlaps with the ground truth box. CIoU goes beyond just calculating overlap, as it considers:

- The aspect ratio of the bounding boxes,
- The distance between the centers of the predicted and the ground truth boxes,
- Size consistency.

*2.5.4.* *AdamW:*

AdamW is an optimization algorithm that modifies the traditional Adam optimizer by decoupling weight decay from the gradient update, which helps to improve generalization. This adjustment prevents the weights from growing too large during training, leading to more stable and efficient convergence.

$$AdamW = \theta_{t+1} = \theta_t - \frac{\alpha}{\widehat{v}_t + \epsilon} * \widehat{m}_t - \alpha * weight\_decay * \theta_t \tag{2}$$

Where:

- α is a learning rate
- $\theta_t$ is the parameter at time step t
- $\epsilon$ is a small constant to avoid division by zero
- $\widehat{v}_t$ is the biased second raw moment estimate
- $\widehat{m}_t$ is the biased first-moment estimate
- weight_decay is the weight decay term [15].

## 2.6. Preprocessing Images for Inference

To improve the model's inference by feeding it with less noisy and more consistent images, several preprocessing techniques were applied.

### 2.6.1. Grayscale Conversion

This technique reduces an image's complexity by removing color data and preserving only the intensity values [16]. It helps to simplify images, mainly when color is not a critical model feature.

### 2.6.2. Gaussian Blur

Gaussian Blur is a technique to reduce image noise and detail by convolving the image with the gaussian function [16]. It helps to smooth out image areas, making it easier for the edge-detection algorithms to perform effectively.

### 2.6.3. Canny Edge Detection

This technique marks the boundaries of objects within an image by identifying and highlighting sharp changes in intensity[18]. It begins with Gaussian smoothing to reduce noise, followed by gradient calculation and edge detection.

### 2.6.4. Cropping and Perspective Transformation

Cropping removes unwanted parts of the image, thus improving the models' ability to focus on the region of interest (ROI). Perspective transformation adjusts the spatial orientation of the image, making it more suitable for analysis.

## 2.7. Application Development

### 2.7.1. Frontend App Development

The application Mero Nagarikta was built and deployed using the Flutter framework, which was built on Dart technology. Flutter allows the project to develop the application for both iOS and Android using the same codebase. It enables users to select photos of citizenship documents' front and back sides, facilitating interaction with the deployed model.

- **Image Selection and Upload:** The App incorporates the ImagePicker package so that users can choose images from the device storage.

- **User Interaction:** When the images are uploaded, users can click a "Fetch" button to forward them to the backend, displaying the extracted information immediately.

- **Data Extraction and Saving:** Users can modify and save extracted information in the .txt file Format. The application has a separate section at the bottom of the screen showing the work history.

*2.7.2. Backend Development and Backend Hosting*

For the backend, the project used Django REST Framework to build APIs that accept POST requests with images in base64-encoded JSON format. The requested images undergo preprocessing before the trained model analyzes them for regions of interest (ROIs). After detection, PyTesseract performs Optical Character Recognition (OCR) to extract text. The project then obtains accurate and validated data through additional processes, such as using the RapidFuzz library for string matching and performing gender and district corrections for specific fields. The cleaned data is returned in JSON format to ensure efficient and accurate interaction between the App and the backend at runtime.

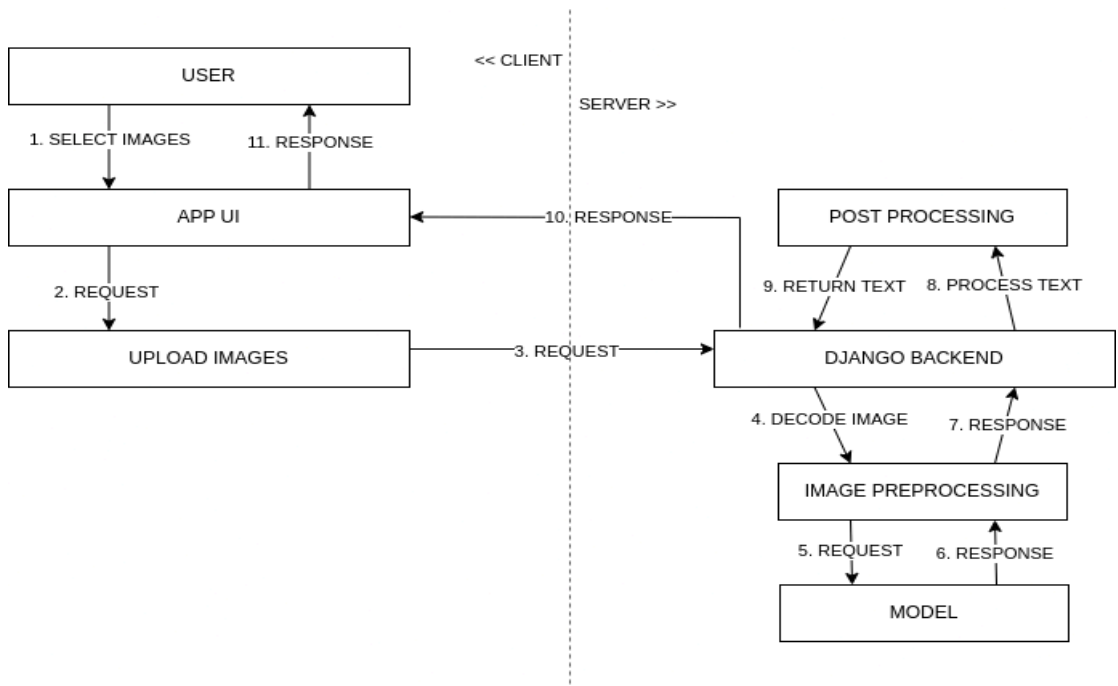

**Figure 3: App Logic.**

## 3. Evaluation Metrics

For the evaluation of the models, several evaluation metrics were utilized:

### 3.1. Intersection over Union (IoU):

IoU measures the overlap between predicted and ground truth bounding boxes, providing a fundamental measure for assessing object localization accuracy.

$$IoU = \frac{Area\ of\ Overlap}{Area\ of\ Union} \quad (3)$$

### 3.2. Accuracy:

Accuracy measures the proportion of correct predictions made by the model compared to the total number of forecasts, calculated as follows:

$$Accuracy = \frac{True\ Positives + True\ Negative}{Total\ number\ of\ sample\ data} \quad (4)$$

### 3.3. Recall:

It measures the ratio of true positives to all actual positives, reflecting the model's ability to detect all instances of a class:

$$Recall = \frac{Total\ Number\ of\ True\ Positives}{True\ Positives + False\ Negatives} \quad (5)$$

### 3.4. Precision:

This metric calculates the ratio of true positives to all optimistic predictions, indicating the model's ability to minimize false positives:

$$Precision = \frac{Total\ Number\ of\ True\ Positives}{True\ Positives + False\ Positives} \quad (6)$$

### 3.5. F1 Score:

The F1 Score, which is the harmonic mean of precision and recall, provides a comprehensive evaluation of model performance by taking into account both false negatives and false positives:

$$F1\ score = 2\frac{Precision * Recall}{Precision + Recall} \quad (7)$$

### 3.6. Support:

Support helps with performance evaluations and identifies potential class imbalances that can have an impact on model results by indicating the number of instances that belong to a particular class.

## 4. Results And Discussion

This project assessed the effectiveness of the YOLOv8 model in detecting text on Nepali citizenship cards through transfer learning. Both models for the front and back are evaluated on the F1 curve, confusion matrix, and precision and recall curves.

*4.1. Model Performance on Front and Back Models:*

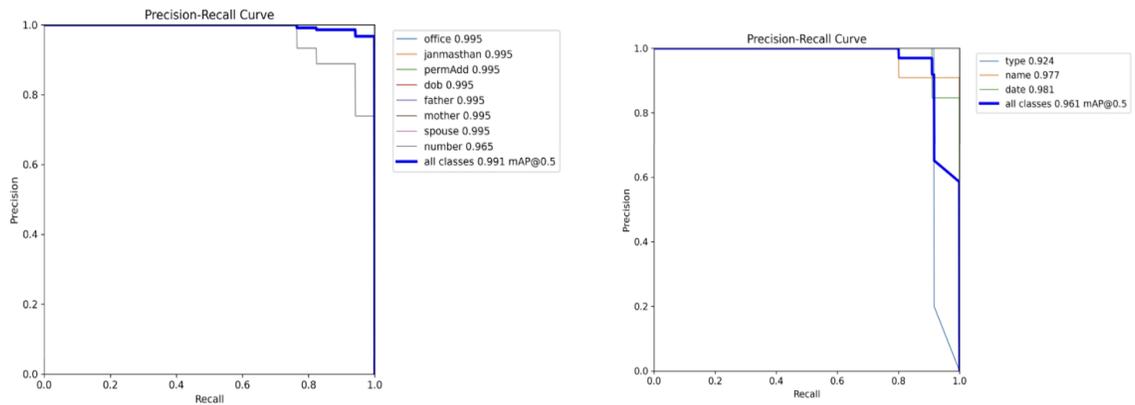

**Figure 4: Precision-Recall Graph of front and back YOLOv8 Model.**

As shown in Figure 4, the Precision-Recall demonstrates the performance metrics for the front and back YOLOv8 models. The Precision-Recall curve represents the performance metrics on both the front and back YOLOv8 models. The front model here is doing pretty great, already achieving a mean Average Precision of 99.1% with a confidence score threshold of 0.5, while the back model trailed poorly at an mAP of 96.1%. Whatever the difference, these results confirmed the capability of this model in handling text recognition across complex settings and provided consistent performance between clean and noisy backgrounds.

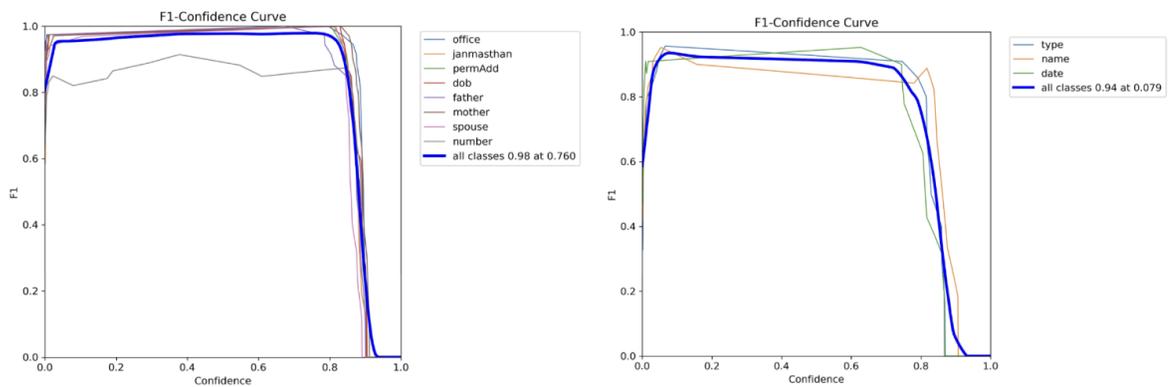

**Figure 5: F1-Confidence Curve of front and back YOLOv8 Model.**

Figure 5 displays the F1-Confidence Curve for both the front and back models. The trade-off between precision and recall at varying confidence thresholds showcases the model's ability to identify different classes with minimal errors. With an F1 score of 0.98, the front model exhibited near-perfect performance. Similarly, the back model scored an F1 score of 0.94, which is a little lower because of smaller typefaces and occlusions (such as stamps). This score emphasizes the robustness of the front model, with only slight issues with the back model.

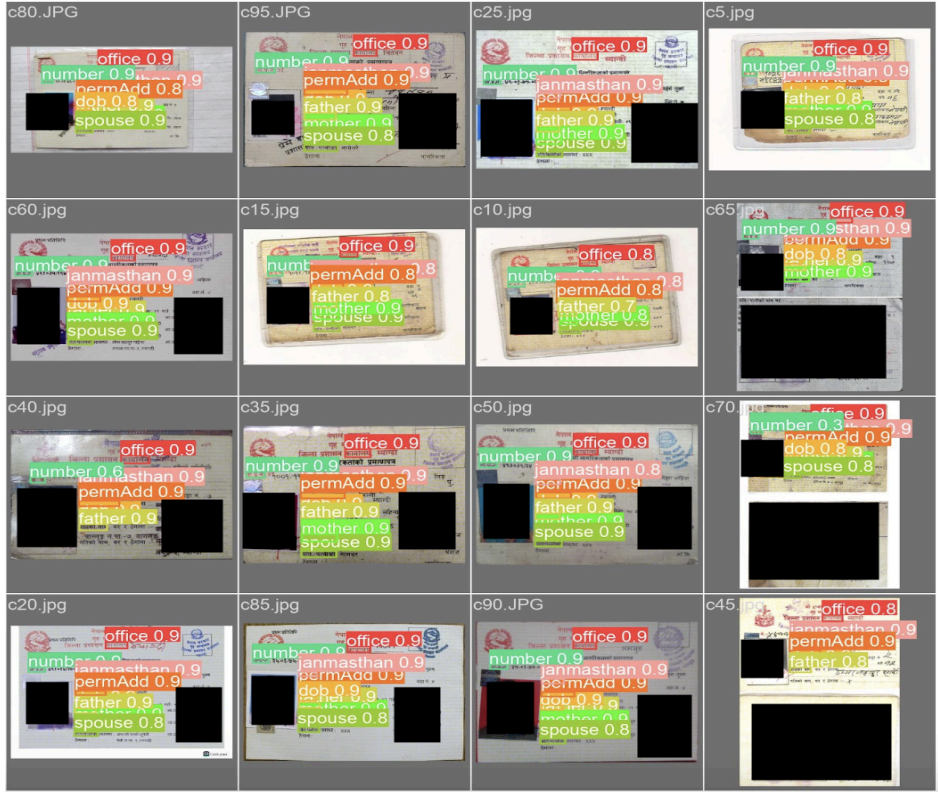

**Figure 6:** Validation Batch 0 for the Front Model. We intentionally added black areas; the model outputs the detection results without them.

### Front Model Performance

The confusion matrix shows how well a model made predictions by comparing the actual true classes to the predicted results, breaking down the counts of correct and incorrect predictions. Figure 7 depicts the confusion matrix of the front model. The front model precisely recognized fields such as name, date of birth, and citizenship number with high recall metrics, showcasing its strong ability to detect these features. Moreover, the confusion matrix highlighted minor challenges faced by the model in recognizing the citizenship number, primarily caused by obstructions from government stamps on the cards.

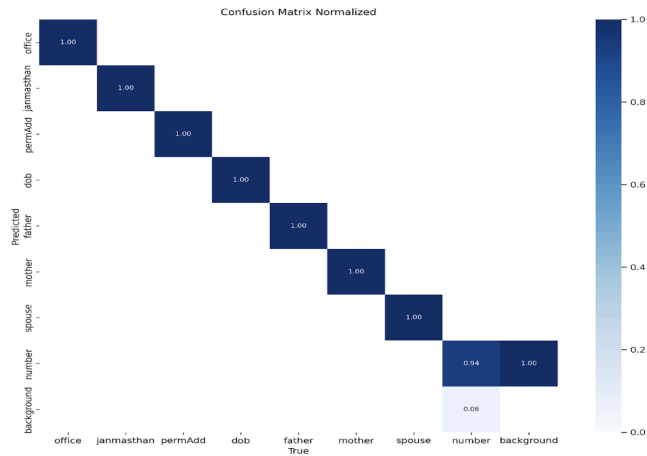

**Figure 7:** Confusion Matrix for Front Model.

**Back Model Performance**

Figure 8, is the confusion matrix for the back model. Although the classes to be detected were fewer for the back model, its result is slightly less accurate than that of the front model as it faces challenges in correctly detecting the issuing officer's name and date of issue .

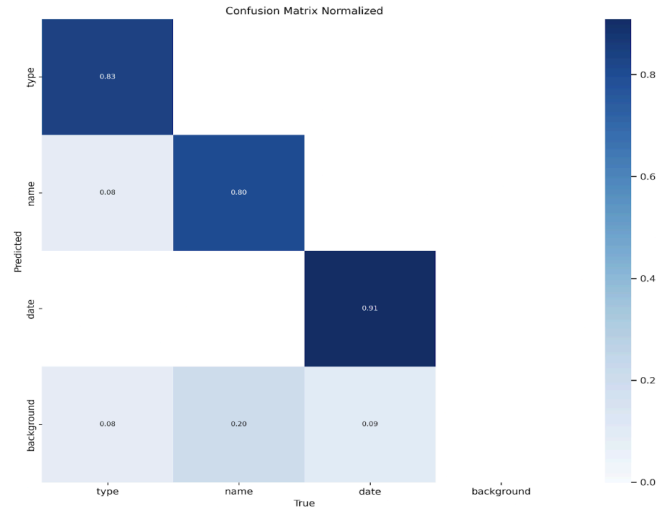

Figure 8: Confusion Matrix for Back Model.

**Post-processing Techniques and Challenges**

After extraction, several post-processing techniques were utilized, including text correction, noise reduction, character replacement, and data organization. These techniques helped to improve the overall accuracy of the extracted data by addressing common OCR errors and misinterpretations. Moreover, the use of language processing techniques including pattern matching and dictionary-based corrections maintained consistency across some fields. For example, the use of sub-tokens in fields like gender and district. Gender-specific terms were also standardized through the identification of linguistic differences to ensure proper categorization. Also, the predefined list of districts was utilized to correct the predicted output for the district field.

Despite all these improvements, some challenges like poor lighting while clicking the photos, variation in text position due to printing errors, and obstructed text from governmental stamps affected the overall performance of the OCR. Highly advanced image preprocessing techniques and models are necessitated to handle complex document layouts and multi-oriented text.

## 5. Conclusion

This project, therefore, identified an effective strategy for parsing text from Nepali citizenship cards by leveraging the power of the deep learning model YOLOv8 in detecting texts and optimized PyTesseract OCR in recognizing texts from the cards. Finally, it has turned out with very impressive performance, which is a mean Average Precision of 99.1% for the front and 96.1% for the back of the cards. It is in this challenging image environment that our preprocessing steps actually proved crucial, such as grayscale conversion and smoothing of the image, which are very important for enhancing the accuracy of OCR. Mero Nagarikta is a mobile application developed as part of the solution-a keen automation approach to extract vital citizenship data by providing a user-friendly interface for real-time text recognition.

In the future, we would like to investigate the inclusion of advanced transformer models that refine the extracted text even further to increase accuracy and precision in information capture. This could then form the basis for future work in significantly raising the overall effectiveness of the system and supporting long-term research in multilingual text processing.

## Conflicts of Interest

The authors declare that there are no conflicts of interest related to the publication of this paper.

## Acknowledgment

The authors express their gratitude to their supervisor, Er. Nabin Lamichhane, for his invaluable guidance and support. They would also like to acknowledge that all authors contributed equally to this work.